\documentclass[preprint]{elsarticle}
\usepackage{xpatch}
\usepackage{array}
\newcolumntype{P}[1]{>{\centering\arraybackslash}m{#1}}
\usepackage[table]{xcolor} 

\usepackage{booktabs} 
\usepackage{array} 
\usepackage{adjustbox}
\usepackage{multirow}
\usepackage{amsmath,amssymb,amsfonts}
\usepackage{algorithm}
\usepackage{algpseudocode}
\usepackage{graphicx}
\usepackage{textcomp}
\usepackage{xcolor}
\usepackage[flushleft]{threeparttable}
\usepackage{enumitem}
\usepackage{{booktabs}}
\usepackage[bottom]{footmisc}
\usepackage{longtable}
\usepackage{caption}
\usepackage{subcaption} 
\usepackage{afterpage}
\usepackage[normalem]{ulem}

    \usepackage{url}
\usepackage[hidelinks]{hyperref}
\xpatchcmd{\State}{\algorithmicend\ \algorithmicfor}{\algorithmicend}{}{}

\usepackage{lineno,hyperref}
\usepackage{float} 

\modulolinenumbers[5]

\begin{document}

\begin{frontmatter}

\title{Multi-Modal Zero-Shot Prediction of Color Trajectories in Food Drying}





\author[UIUCMechSE]{Shichen Li}
\ead{shichen8@illinois.edu}

\author[UIUCMechSE]{Ahmadreza Eslaminia}
\ead{ae15@illinois.edu}

\author[UMich,UIUCMechSE]{Chenhui Shao\corref{mycorrespondingauthor}}
\ead{chshao@umich.edu}

\address[UIUCMechSE]{Department of Mechanical Science and Engineering, University of Illinois at Urbana-Champaign, Urbana, IL, USA}

\address[UMich]{Department of Mechanical Engineering, University of Michigan, Ann Arbor, MI, USA}

\cortext[mycorrespondingauthor]{Corresponding author}

\begin{abstract}
Food drying is widely used to reduce moisture content, ensure safety, and extend shelf life. Color evolution of food samples is an important indicator of product quality in food drying. Although existing studies have examined color changes under different drying conditions, current approaches primarily rely on low-dimensional color features and cannot fully capture the complex, dynamic color trajectories of food samples. Moreover, existing modeling approaches lack the ability to generalize to unseen process conditions. To address these limitations, we develop a novel multi-modal color-trajectory prediction method that integrates high-dimensional temporal color information with drying process parameters to enable accurate and data-efficient color trajectory prediction. Under unseen drying conditions, the model attains RMSEs of 2.12 for cookie drying and 1.29 for apple drying, reducing errors by over 90\% compared with baseline models. These experimental results demonstrate the model’s superior accuracy, robustness, and broad applicability.

\end{abstract}


\end{frontmatter}


\section{Introduction}\label{sec:introduction}

As a fundamental operation in industrial food processing, drying enables long-term preservation, enhances texture and flavor, and facilitates storage and transportation~\cite{oliveira2016influence}. However, food drying is a highly complex process~\cite{li2020novel}. External process parameters such as drying temperature and air velocity, along with internal factors like product size, shape, and composition, contribute significantly to its nonlinear dynamics~\cite{li2025uncertainty}.
Due to these complexities, observing only the initial and final states of the drying process fails to capture its full dynamic behavior. Instead, continuous tracking of attribute changes throughout the drying period provides deeper insights into process evolution and supports real-time quality control~\cite{tian2023weldmon}. Such attributes monitoring enables more effective process adjustment and optimization~\cite{iheonye2023monitoring}.

Existing studies have primarily tracked food drying kinetics, such as moisture loss, weight reduction, and shape deformation~\cite{inyang2018kinetic, delfiya2022drying, nasri2018effects}. These efforts have advanced the understanding of drying behavior under fixed process parameters and controlled sample conditions. Among the monitored attributes, color change is particularly valuable due to its pronounced variability during drying~\cite{li2020novel}. Moreover, the final color of dried products strongly influences consumer perception and is closely associated with key quality indicators such as texture and moisture content~\cite{perera2005selected}. Therefore, the trajectory of color change serves as an important indicator for process quality control and early detection of abnormal drying behavior~\cite{salehi2018modeling}.

There is a lack of effective methods for tracking food attributes in food drying. Manual measurements often disrupt sample integrity and make the process irreversible~\cite{zambrano2019assessment}. As a result, most studies rely on discrete post-drying measurements at intermediate stages, repeated under identical conditions~\cite{boateng2021effect}. This approach is inefficient and fails to account for natural variability among samples. While computer vision (CV)-based methods enable non-invasive, continuous monitoring of color change~\cite{iheonye2023monitoring}, they remain limited to passive observation and lack predictive capabilities, making it difficult to detect abnormal drying behaviors in real-time. Moreover, existing methods are typically tailored to specific process conditions and generalize poorly across varying parameters and sample characteristics~\cite{gaikwad2022effect}. Their reliance on fixed formulations or empirical models further limits their applicability to more complex scenarios in industrial process modeling and control.

Consequently, it is essential to develop predictive methods that forecast color-change trajectories. Machine learning (ML) techniques offer a promising solution due to their ability to model complex, nonlinear dynamics from historical drying data~\cite{meng2023explainable}. However, given the vast parameter space of industrial food drying processes, compiling exhaustive training datasets to cover all possible conditions is impractical~\cite{djakovic2024review}. To make the most efficient use of limited, cost-intensive food drying data, it is desirable to generalize from a small set of known conditions to a wide range of unseen ones~\cite{meng2024meta}. This motivates the development of ML models that can perform zero-shot trajectory prediction while accounting for variability in sample characteristics~\cite{li2025multi}.

To the best of our knowledge, no existing studies have systematically addressed the prediction of attribute trajectories under zero-shot learning for unseen process conditions. Predicting color change during drying presents distinct challenges due to its inherently non-monotonic behavior, measurement noise, limited data availability, and significant sample-to-sample variability~\cite{you2024prediction}. The zero-shot setting adds further difficulty, as training and evaluation datasets are often non-overlapping and unevenly distributed~\cite{pourpanah2022review}. Addressing this problem requires a framework that can extract generalizable patterns from limited data while remaining robust to both process variability and sparse observations.

In this work, we propose a novel multi-modal data-driven modeling approach for zero-shot color-change trajectories prediction during food drying. Our method is able to learn parameters of component functions that mathematically represent color evolution over time. To improve robustness and generalizability, we apply discrete cosine transform (DCT) preprocessing to reduce measurement noise and capture the dominant trajectory trends. We further enhance predictive performance through two strategies: multi-modal data fusion, and similarity-informed selection of training datasets. The proposed method is evaluated against a baseline moment-by-moment trajectory prediction model using separate, non-overlapping training and evaluation datasets. We validate the approach on two distinctive and representative case studies, cookie drying and apple drying, to demonstrate accurate trajectory prediction across varying color dynamics. Experimental results show the model achieves an RMSE of 2.12 for cookie drying and 1.29 for apple drying under unseen conditions, substantially outperforming baseline models with RMSEs of 31.36 and 10.16, respectively. These results highlight the potential of our method for real-time, data-driven quality monitoring in food drying applications.

The remainder of this paper is structured as follows: Section 2 reviews related work. Section 3 details our proposed methodology, baseline method, and strategies to enhance predictive accuracy. Section 4 presents case studies with experimental results. Finally, Section 5 concludes the paper and suggests future directions.

\section{Related Work}\label{sec:related_work}

\subsection{Zero-Shot Learning}

Zero-shot learning refers to the ability of an ML model to make predictions on tasks or data it has never encountered during training~\cite{ye2022zerogen}. Rather than relying on traditional i.i.d. (independent and identically distributed) assumptions, zero-shot learning focuses on generalizing to novel conditions by leveraging prior knowledge to bridge the gap between seen and unseen cases~\cite{eslaminia2024federated}. It has been widely applied in areas such as CV~\cite{zheng2022semantic}, natural language processing~\cite{sivarajkumar2023healthprompt}, and robotics~\cite{jang2022bc}, where model deployment often involves conditions that differ from the training environment.

\begin{figure}[!t]
\centerline{\includegraphics[width=\columnwidth]{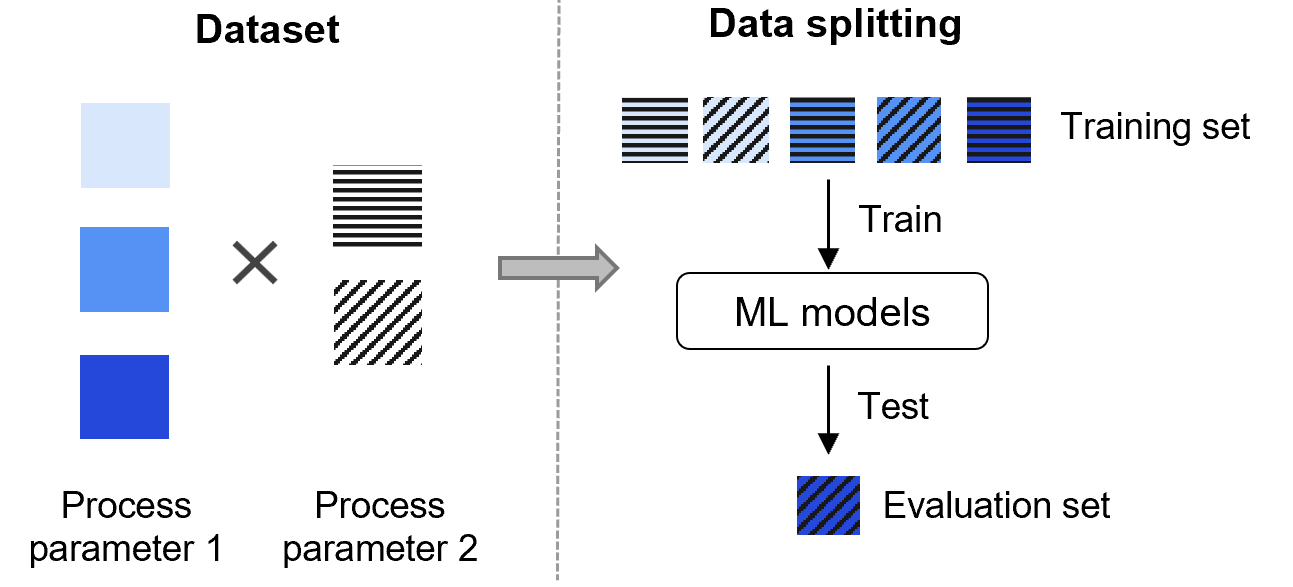}}
\caption{Zero-shot learning-based data splitting strategy.}
\label{fig1}
\end{figure}

In our context, zero-shot learning refers to the non-overlapping split between training and evaluation datasets based on controllable process parameters in food drying (e.g., drying temperature and air velocity), the strategy is illustrated in Fig.~\ref{fig1}. Each unique combination of process parameters defines a distinct drying condition. During training, the model sees only a subset of these combinations and is tested on completely unseen ones. Specifically, let $\mathbf{x} = [x_1, x_2] \in \mathcal{X}$ represent the process parameters, and $\mathbf{y}(t) \in \mathbb{R}^n$ denote the corresponding color-change trajectory over time. The goal is to learn a predictive function $f_\theta : \mathcal{X} \rightarrow \mathbb{R}^n$, such that:
\begin{equation}
    \hat{\mathbf{y}} = f_\theta(\mathbf{x}).\label{eq}
\end{equation}

Here, $\mathcal{D}_{\text{train}} = \{(\mathbf{x}_i, \mathbf{y}_i)\}_{i=1}^N$ consists of training samples from a subset of the process space $\mathcal{X}_{\text{train}}$, while the evaluation set $\mathcal{D}_{\text{eval}}$ includes samples from unseen conditions, i.e., $\mathcal{X}_{\text{eval}} = \mathcal{X} \setminus \mathcal{X}_{\text{train}}$. To date, existing food drying prediction models have not adopted a zero-shot learning framework.

\subsection{Trajectory Prediction}

A trajectory refers to the temporal evolution of a specific attribute or variable over time~\cite{spaccapietra2008conceptual}. Trajectory prediction typically involves learning patterns from past data to forecast future values, thereby capturing the dynamics of sequential or time-dependent processes~\cite{koolwal2020comprehensive}. This approach has been widely applied in domains such as robotic motion planning~\cite{chen2024differentiable}, human activity forecasting~\cite{kothari2021human}, and traffic flow prediction~\cite{li2021traffic}, where the objective is to anticipate future states based on historical trends and contextual inputs.

In the context of food drying, trajectory modeling has traditionally focused on tracking drying kinetics through discrete measurements, which employs curve fitting or estimating drying rates under fixed process parameters~\cite{ruiz2007analytical, adeyeye2019overview}. However, there has been no work on trajectory prediction in food drying applications. Given the inherent noise and limited resolution of food drying datasets, our approach departs from moment-by-moment prediction and instead focuses on learning the overall trends of attribute evolution using a data-driven modeling framework~\cite{champion2019data}.

\subsection{Data-Driven Modeling}

Data-driven modeling refers to the process of learning underlying patterns, structures, or functional relationships directly from observed data, often without relying on explicit first-principles equations~\cite{montans2019data}. This approach leverages ML, statistical inference, and signal processing techniques to model complex systems where analytical modeling is difficult or impractical~\cite{jia2024hybrid}. Data-driven methods have been widely adopted across domains such as financial forecasting~\cite{al2025financial}, anomaly detection~\cite{eslaminia2025fdm}, and biomedical signal analysis~\cite{dai2013model}, where the systems are governed by nonlinear, high-dimensional, or partially observable dynamics.

In more specialized settings, data-driven modeling can involve the discovery of latent coordinates, component functions, or parametric representations that compactly capture the evolution of a process over time~\cite{champion2019data}. These methods aim to express complex trajectories using a small set of interpretable components or learned representations, enabling efficient learning, generalization, and reconstruction~\cite{liu2022survey}. In this work, we adopt such a data-driven approach to model the temporal evolution of color during food drying, treating it as a function influenced by both external process parameters and internal sample characteristics. Methodological details are provided in Section~\ref{sec:methodology}.



\section{Methodology}
\label{sec:methodology}

This section presents the data-driven trajectory prediction methodology for food drying. Section 3.1 introduces the model architecture used for trajectory prediction. Sections 3.2 and 3.3 detail three strategies to enhance prediction accuracy: Section 3.2 explores multi-modal data fusion by incorporating the initial image state; Section 3.3 investigates filtering the training set to include only similar drying cases. Finally, Section 3.4 presents a moment-by-moment prediction approach as a suboptimal baseline for comparison.

\subsection{Model Architecture}

\begin{figure*}[!t]
\centerline{\includegraphics[width=0.99\textwidth]{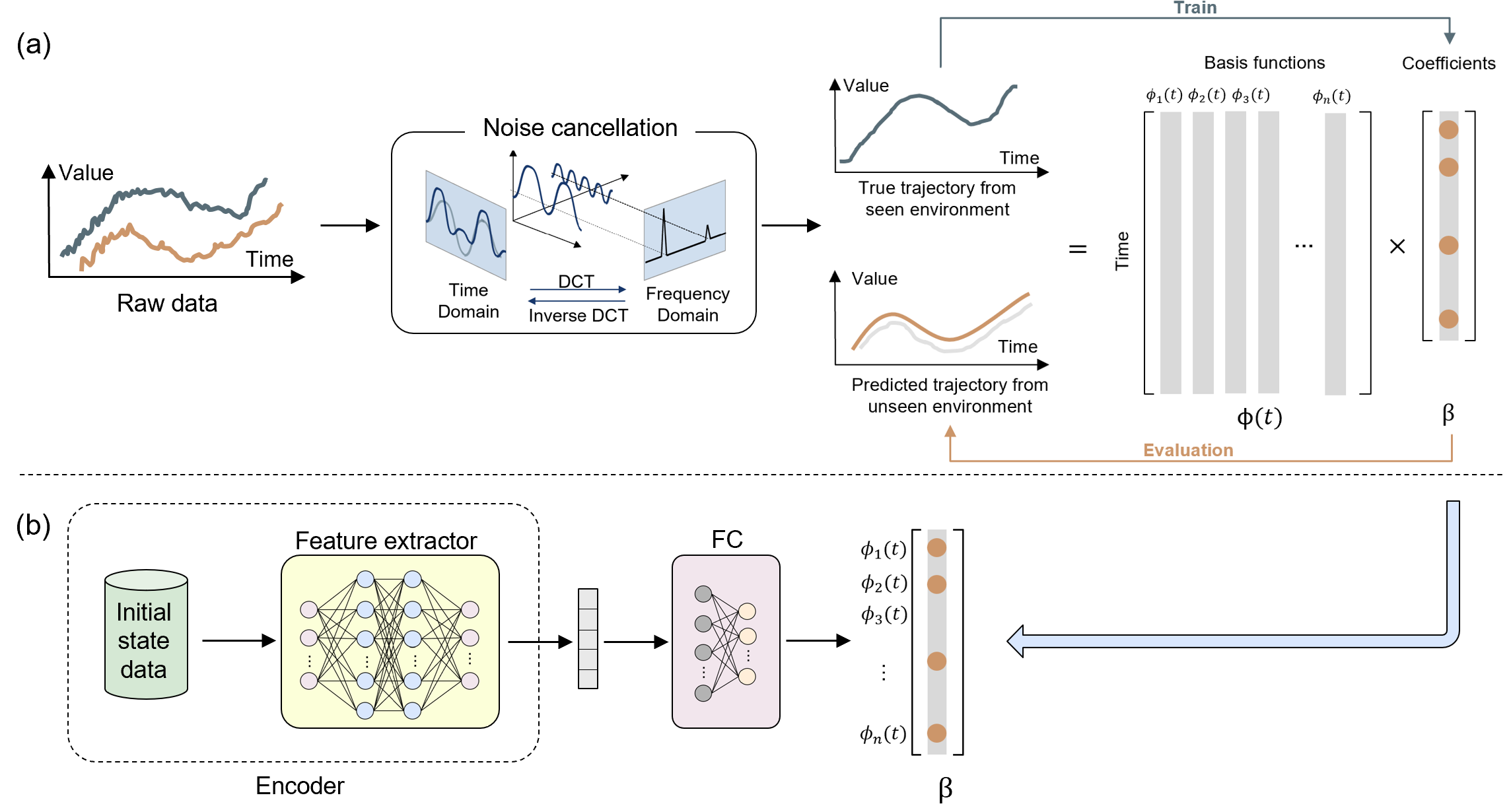}}
\caption{Overall model architecture of data-driven trajectory prediction.}
\label{fig2}
\end{figure*}

We propose a data-driven modeling method for predicting color-change trajectories in food drying by learning the coefficients of a set of component functions that together form a compact mathematical representation of the trajectory. The overall model architecture is illustrated in Fig.\ref{fig2} .

Due to the noise inherent in real-world food drying datasets, the raw trajectory data is first preprocessed using the DCT. DCT is a widely used signal processing technique that transforms a time-domain signal into a sum of cosine functions oscillating at different frequencies~\cite{ahmed2006discrete}. This converts the trajectory from the time domain to the frequency domain, allowing for better identification and suppression of high-frequency noise~\cite{robertson2005dct}. An inverse DCT is then applied to reconstruct a denoised version of the trajectory in the time domain, preserving the primary trend while eliminating unwanted fluctuations. This preprocessing step improves learning stability and reduces the effect of measurement artifacts.

We formulate trajectory prediction as the learning of a smooth function of time, represented by a weighted combination of predefined component functions $\{\phi_1(t), \phi_2(t), \ldots, \phi_n(t)\}$. The predicted trajectory $\hat{y}(t)$ is expressed as:

\begin{equation}
    \hat{y}(t) = \sum_{i=1}^{n} \beta_i \phi_i(t) = \boldsymbol{\phi}(t)^\top \boldsymbol{\beta},\label{eq}
\end{equation}
where $\boldsymbol{\phi}(t) \in \mathbb{R}^n$ is the vector of component functions evaluated at time $t$, and $\boldsymbol{\beta} \in \mathbb{R}^n$ is the coefficient vector to be learned.

The final set of component functions used in this work consists of a combination of polynomial, trigonometric, logarithmic terms, and their products. These functions were chosen for their ability to capture the diversity of drying trajectories. During training, different model configurations automatically select and weight subsets of these functions, with the estimation of coefficients being the primary challenge and focus of this work. The complete set of functions is listed below:


\begin{equation}
\left\{
\begin{aligned}
&1,\, t,\, t^2,\, \frac{1}{t + 1},\, \sin(2\pi t),\, \cos(2\pi t),\,  \\
&t \sin(2\pi t),\, t \cos(2\pi t),\, \log(t)
\end{aligned}
\right\}
\label{eq}
\end{equation}

To predict the coefficient vector $\boldsymbol{\beta}$, we use a neural network architecture illustrated in Fig.~\ref{fig2}(b). The input to the network is the initial state data, which may include process parameters and sample-specific features. This input is passed through an encoder network that performs feature extraction to produce a low-dimensional latent embedding. The architecture of the encoder is adaptable depending on the input format. 

The extracted features are then passed through a fully connected (FC) layer to generate the coefficient vector $\boldsymbol{\beta}$, which defines the trajectory. The reconstructed trajectory is compared to the ground truth at each time step using the summed mean squared error (MSE) loss:

\begin{equation}
\mathcal{L} = \frac{1}{T} \sum_{t=1}^{T} \left( y(t) - \hat{y}(t) \right)^2,\label{eq}
\end{equation}
where $y(t)$ is the observed trajectory value at time $t$, and $T$ is the total number of time steps.

Training proceeds until the model either reaches a predefined 200 number of epochs or achieves convergence based on the validation loss. Importantly, the training and evaluation datasets are non-overlapping in process parameters, establishing a zero-shot learning setting, where the model must generalize to previously unseen drying conditions. The overall pseudocode is shown in Algorithm 1.

\begin{algorithm}[H]
\caption{\textsc{Data-Driven Trajectory Prediction}}
\label{alg:zero_shot}
\begin{algorithmic}[1]
\Require Dataset $\mathcal{D} = \{(x_i, y_i(t))\}$, component functions $\boldsymbol{\phi}(t)$
\Ensure Trained model, predicted trajectories $\hat{y}(t)$

\Procedure{TrajectoryPrediction}{$\mathcal{D}, \boldsymbol{\phi}(t)$}
    
    \State // \textbf{Preprocessing}
    \ForAll{$(x_i, y_i(t)) \in \mathcal{D}$}
        \State Apply DCT to $y_i(t) \rightarrow Y_{\text{freq}}$
        \State Suppress high-frequency components in $Y_{\text{freq}}$
        \State Apply inverse DCT to obtain denoised $\tilde{y}_i(t)$
    \EndFor

    \State // \textbf{Split dataset into training and evaluation sets}
    \State Split $\mathcal{D}$ into $\mathcal{D}_{\text{train}}$ and $\mathcal{D}_{\text{eval}}$
    
    \State // \textbf{Training}
    \ForAll{$(x_i, \tilde{y}_i(t)) \in \mathcal{D}_{\text{train}}$}
        \State Encode $x_i$ with neural encoder $\rightarrow$ feature vector $z_i$
        \State Predict coefficients: $\boldsymbol{\beta}_i = \text{FC}(z_i)$
        \State Reconstruct trajectory: $\hat{y}_i(t) = \boldsymbol{\phi}(t)^\top \boldsymbol{\beta}_i$
        \State Compute MSE loss $\mathcal{L}$ between $\hat{y}_i(t)$ and $\tilde{y}_i(t)$
        \State Update model parameters to minimize $\mathcal{L}$
    \EndFor

    \State // \textbf{Evaluation (Zero-Shot)}
    \ForAll{$(x_j, y_j(t)) \in \mathcal{D}_{\text{eval}}$}
        \State Encode $x_j \rightarrow z_j$
        \State Predict coefficients: $\hat{\boldsymbol{\beta}}_j = \text{FC}(z_j)$
        \State Reconstruct: $\hat{y}_j(t) = \boldsymbol{\phi}(t)^\top \hat{\boldsymbol{\beta}}_j$
        \State Evaluate prediction using MSE
    \EndFor

    \State \Return Trained model, predicted $\hat{y}(t)$
\EndProcedure
\end{algorithmic}
\end{algorithm}

\subsection{Multi-Modal Fusion of Initial Stage Data}

To further improve the prediction accuracy of the proposed method, we incorporate a multi-modal fusion strategy within the encoder. In food drying applications, input data typically include process parameters in tabular format that define the external drying conditions. However, even under identical environmental settings, the resulting color-change trajectories often vary significantly due to sample-specific factors such as geometry, surface texture, or composition~\cite{li2025uncertainty}. These internal sample characteristics are typically difficult to quantify or represent using manually defined scalar features.

To capture this hidden variability, we enrich the initial state representation through multi-modal fusion by combining both structured process parameters and unstructured visual information~\cite{li2025multi2}. Specifically, we utilize the first captured image of each food sample before drying begins. This image implicitly encodes valuable information about initial sample conditions that are otherwise hard to measure or define.

\begin{figure}[!t]
\centerline{\includegraphics[width=\columnwidth]{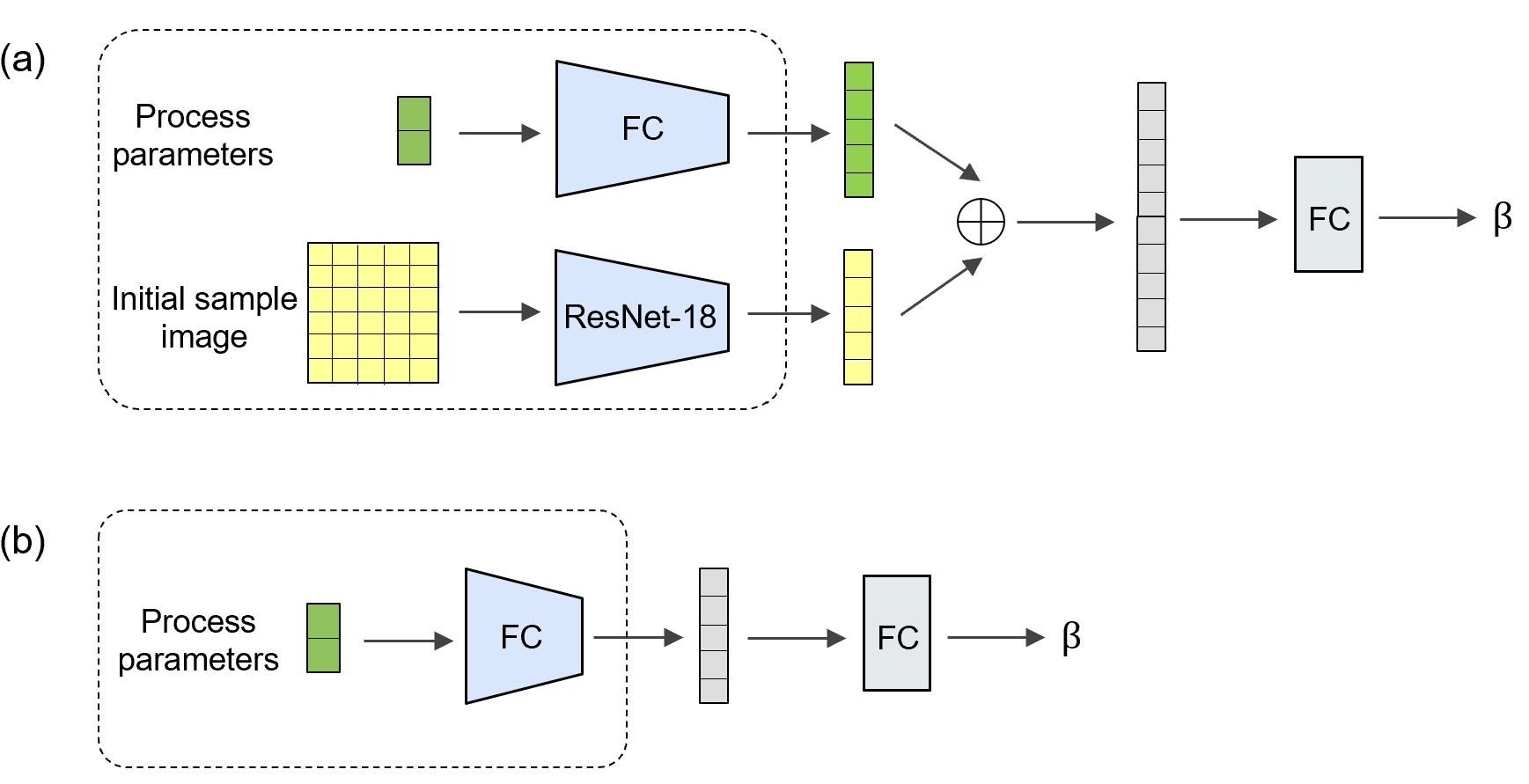}}
\caption{Comparison between (a) multi-modal fusion encoder and (b) tabular-only encoder.}
\label{fig3}
\end{figure}

Fig.~\ref{fig3} presents the comparison between two encoder architectures. Specifically, Fig.~\ref{fig3}(a) shows the multi-modal fusion model. The process parameters are passed through the FC layer to generate a compact feature vector. Simultaneously, the initial image is passed through a pre-trained ResNet-18 to extract visual features. The two feature vectors are then concatenated and fed into an additional FC layer to predict the component function coefficients $\boldsymbol{\beta}$. Fig.~\ref{fig3}(b) illustrates the signle-modality model using only process parameters, which are encoded through sequential FC layers for coefficient prediction.

This fusion strategy allows the model to leverage raw visual input without requiring handcrafted feature engineering or prior image preprocessing into tabular format. Our previous studies have shown that incorporating multi-modal input substantially improves food drying prediction accuracy, especially under high-variance conditions introduced by sample-level heterogeneity~\cite{li2025multi2}.

\subsection{Similarity-Informed Selection of Training Datasets}

In food drying applications, datasets are often limited in both size and quality, frequently containing measurement noise and manual errors. Additionally, since our objective is to predict trajectories in a zero-shot setting, the data distribution is inherently non-i.i.d., violating assumptions of many standard ML frameworks and complicating generalization.

Under these constraints, simply increasing the amount of training data does not always lead to better performance. Instead, we focus on selecting a subset of training data that is most relevant to the evaluation condition, thereby enabling more targeted and informative learning. 

We hypothesize that samples collected under similar drying conditions share latent feature patterns that are more informative for predicting trajectories in related but unseen settings. This motivates a similar-condition training strategy, where we select training samples that share one of the same process parameters as the evaluation set. 

\begin{figure}[!t]
\centerline{\includegraphics[width=\columnwidth]{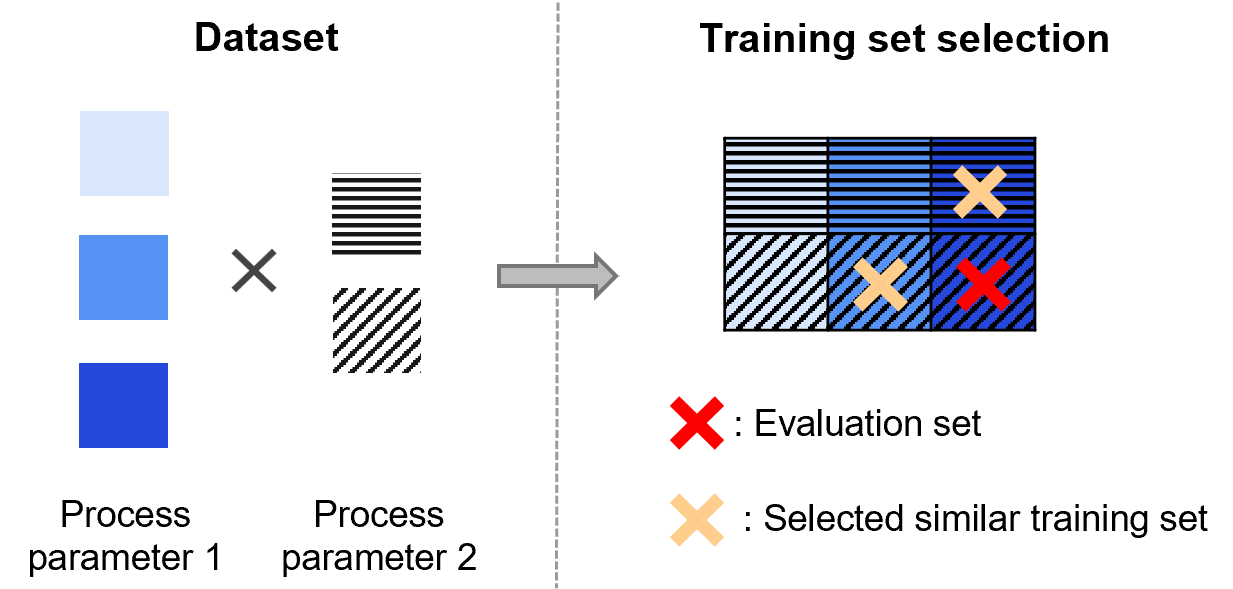}}
\caption{Similarity-informed selection of training datasets.}
\label{fig4}
\end{figure}

As illustrated in Fig.~\ref{fig4}, each box represents a unique combination of process parameters. Red markers indicate evaluation conditions, while yellow markers represent the selected training data with similar conditions. This targeted selection approach is inspired by transfer learning, where knowledge from related similar domains is leveraged to improve performance under data scarcity.

Rather than training on the entire dataset indiscriminately, this method prioritizes informative samples that are more likely to support accurate trajectory prediction under novel process conditions.

\subsection{Baseline Model}

We compare the effectiveness of our method against a moment-by-moment prediction baseline. In this baseline, the model sequentially predicts the next time point of the trajectory based on current and past observations, without explicitly modeling the global structure of the curve.

To implement this, we use a Long Short-Term Memory (LSTM) network, which is well-suited for time-series forecasting due to its ability to capture temporal dependencies~\cite{yu2019review}. The LSTM is trained to predict the color change value at the next timestamp given values from previous 5 timestamps previous five values, and iteratively rolls out the full trajectory by feeding its own predictions back into the input.

This baseline serves as a contrast to our coefficient-based formulation, as it models the trajectory in an autoregressive fashion without imposing a structured functional form. We investigate the two approaches in terms of smoothness, robustness to noise, and generalization to unseen process conditions.

\section{Case Studies}
\label{sec:cases}

In this section, we apply our method on two case studies: apple drying and cookie drying, to demonstrate the effectiveness of the proposed method, compared with the baseline model and try on two further improvements directions. Specifically, Section 4.1 is color trajectory prediction in cookie drying, and Section 4.2 is color trajectory prediction in apple drying. 

\subsection{Color Trajectory Prediction in Cookie Drying}

\subsubsection{Experiments}

Sugar cookie drying is a high-temperature process in which the surface color changes rapidly and significantly, reflecting the extent of drying and indicating potential over-drying or under-drying~\cite{farzad2021drying}. This makes it well-suited for studies on color trajectory prediction.

We perform controlled drying experiments to collect cookie color trajectory data under varying process conditions (temperature and air velocity). The experiments use a Nuwave Smart Oven in bake mode, with both top and bottom heating elements and a convection fan. We test four temperature levels (350, 375, 385, 400${}^\circ\mathrm{F}$) and two fan speeds (1000 and 3000 RPM), resulting in eight drying conditions. Each condition is repeated three times with three cookies per batch, yielding 24 experiments and 72 samples. The drying duration under each process condition is fixed at 12 minutes.

We use Pillsbury sugar cookie dough as the raw sample. Each piece weighs approximately 18.5 g and is formed into a round shape with a diameter of 60–70 mm and a thickness of 6–7mm. After drying, the cookies reduce to 15–16g in weight and expand to a diameter of 80–90mm and a thickness of 8–10mm.

\begin{figure}[!t]
\centerline{\includegraphics[width=0.99\columnwidth]{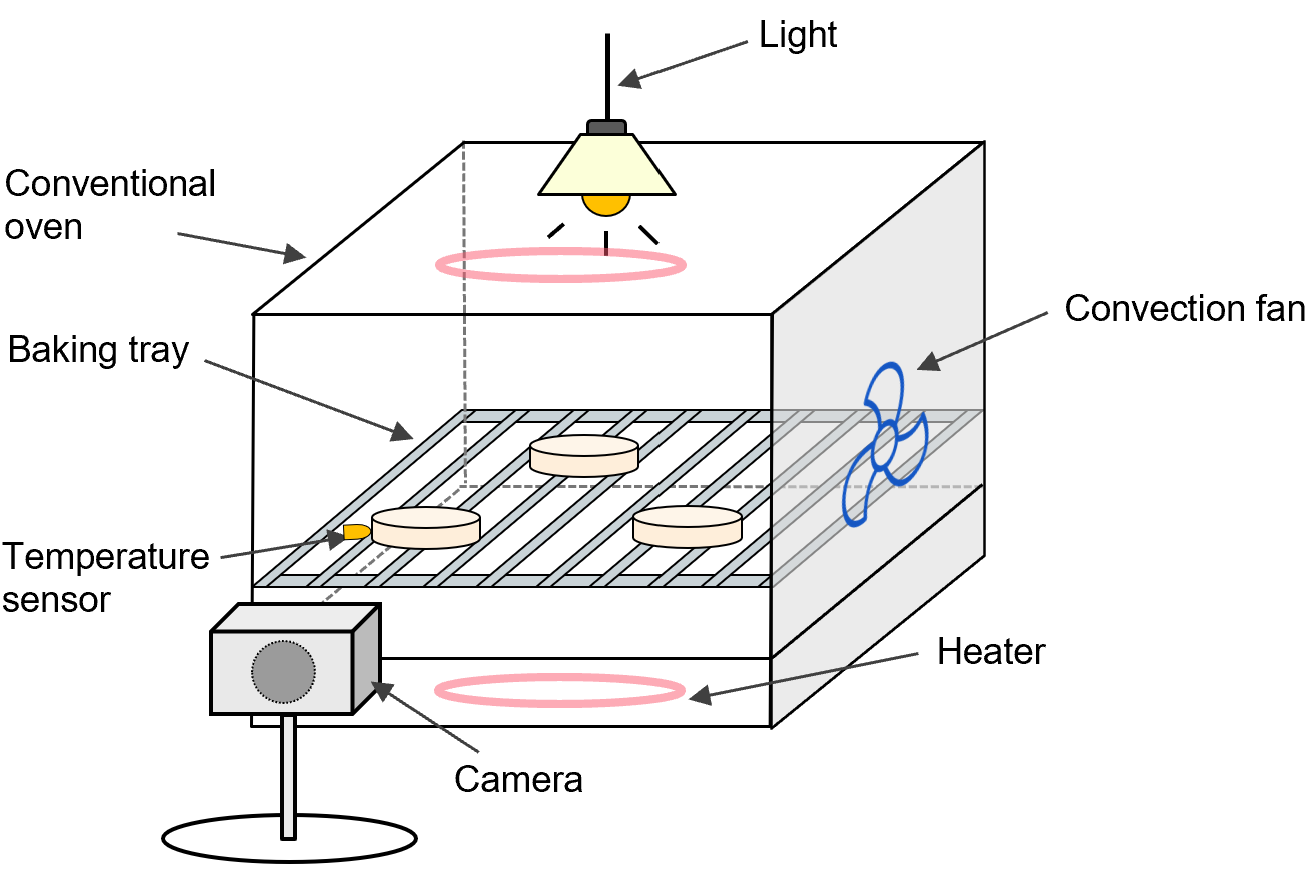}}
\caption{Experimental setup for cookie drying.}
\label{fig5}
\end{figure}

Fig.~\ref{fig5} illustrates the experimental setup. During the cookie drying process, a camera records one frame every 10 seconds through the oven’s glass door, aided by a 5000K LED light to maintain stable illumination. Although minor reflection noise is introduced, the setup consistently captures real-time color change for three cookies per frame. 

\subsubsection{Data Preparation}

We utilize image data to extract the average color values of cookie surfaces over time, enabling efficient in-situ monitoring of color change without the need for repeated experiments. To segment individual cookies in each frame, we apply the Segment Anything Model (SAM), which provides robust and automated image segmentation across time series data~\cite{kirillov2023segment}. As illustrated in Fig.~\ref{fig5}, SAM takes each raw image as input, processes it through an image encoder, and generates instance-specific masks using prompt-based decoding. The resulting masks accurately isolate each cookie sample, allowing valid per-cookie color measurements to be extracted throughout the drying process.

\begin{figure}[!t]
\centerline{\includegraphics[width=\columnwidth]{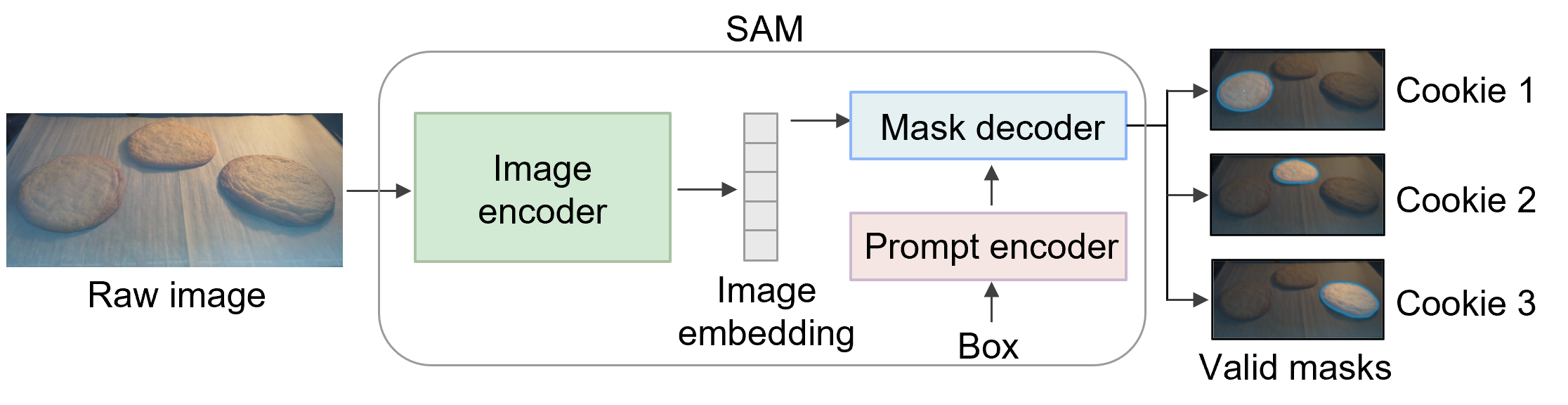}}
\caption{SAM for cookie drying images.}
\label{fig6}
\end{figure}

\begin{figure}[!t]
\centerline{\includegraphics[width=\columnwidth]{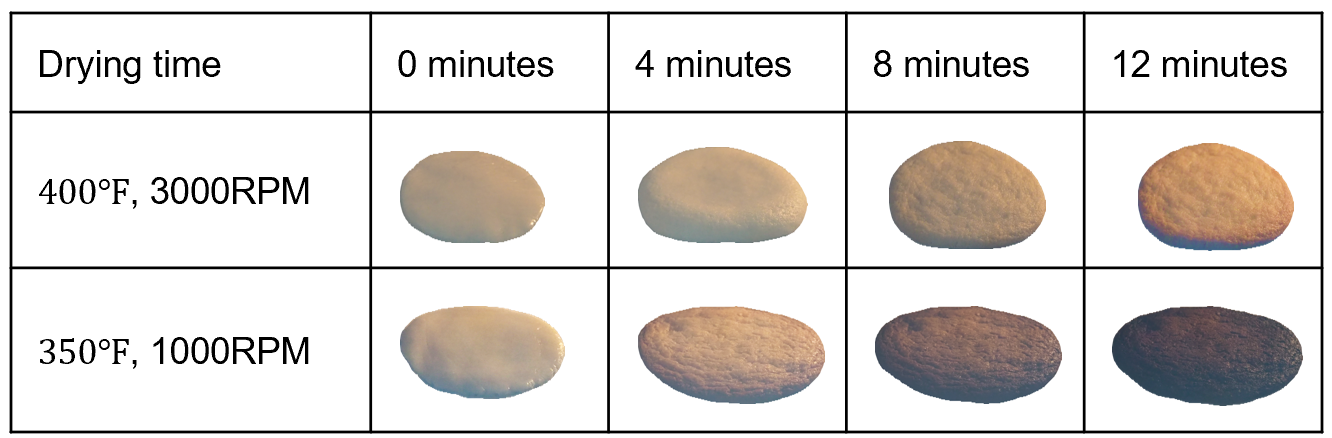}}
\caption{Masked cookie samples under 350${}^{\circ}\mathrm{F}$, 1000RPM and 400${}^{\circ}\mathrm{F}$, 3000RPM.}
\label{fig7}
\end{figure}

Fig.~\ref{fig7} shows an example of two time-series masked cookie samples, where clear color differences can be observed both along the duration of each drying process and across different process conditions at the same time point. We extract the pixel-wise average color value from each segmented mask and use it as the trajectory variable over time.

The color change during drying is quantified using the CIE color difference metric, denoted as $\Delta E$, which measures the Euclidean distance between the initial color of the sample and its color at a given time $t$ in the CIELAB color space. It is computed as:

\begin{equation}
\Delta E(t) = \sqrt{[L^*(t) - L_0^*]^2 + [a^*(t) - a_0^*]^2 + [b^*(t) - b_0^*]^2},\label{eq}
\end{equation}
where $L_0^*, a_0^*, b_0^*$ represent the initial lightness and chromaticity values of the sample surface at the beginning of the drying process ($t = 0$), and $L^*(t), a^*(t), b^*(t)$ denote the corresponding values at time $t$. This metric provides a perceptually meaningful measure of the overall color deviation from the initial state and is widely used in food quality evaluation~\cite{falade2012effects}.

\begin{figure}[!t]
\centerline{\includegraphics[width=0.8\columnwidth]{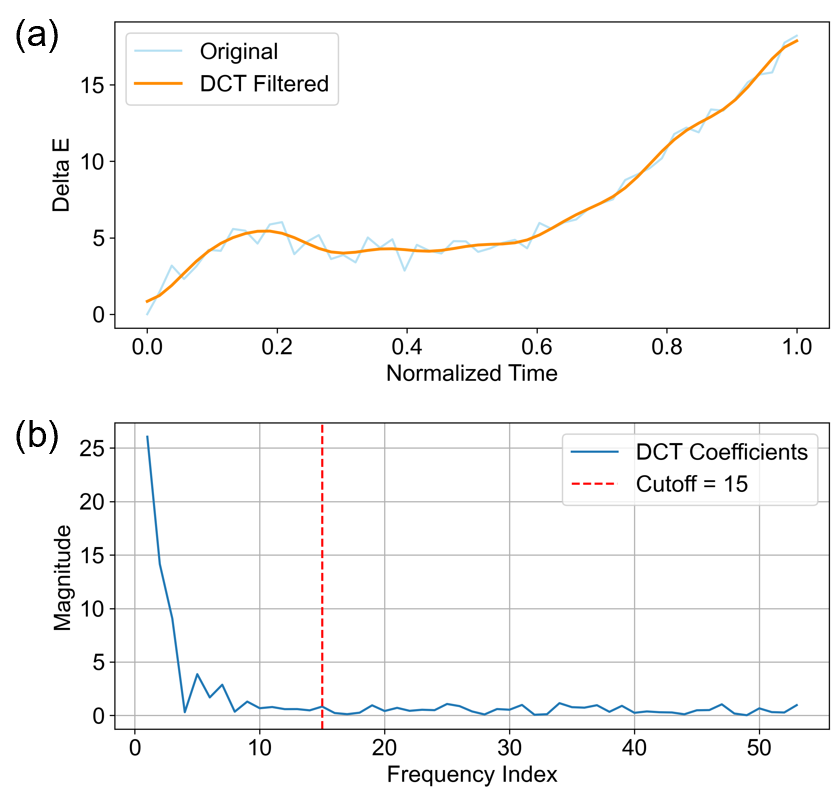}}
\caption{Color trajectory noise cancellation for cookie drying with (a) original and smoothed average trajectory under 400${}^{\circ}\mathrm{F}$, 1000RPM and corresponding (b) DCT coefficient spectrum.}
\label{fig8}
\end{figure}

Fig.~\ref{fig8} presents an example of a color trajectory in terms of $\Delta E$ over normalized drying time. The original trajectory contains small fluctuations, which we attribute to environmental noise such as lighting variations and camera angle instability. To address this, we apply a DCT-based smoothing technique to reduce high-frequency components while preserving the underlying trend. Fig.~\ref{fig8}(b) shows the corresponding DCT coefficient spectrum, where most of the signal energy is concentrated in the low-frequency range. A cutoff threshold is applied at index 15 to remove noise-dominant components. The resulting smoothed trajectory, shown in orange in Fig.~\ref{fig8}(a), is used as the ground truth for model training and evaluation.

\begin{figure}[!t]
\centerline{\includegraphics[width=1.05\columnwidth]{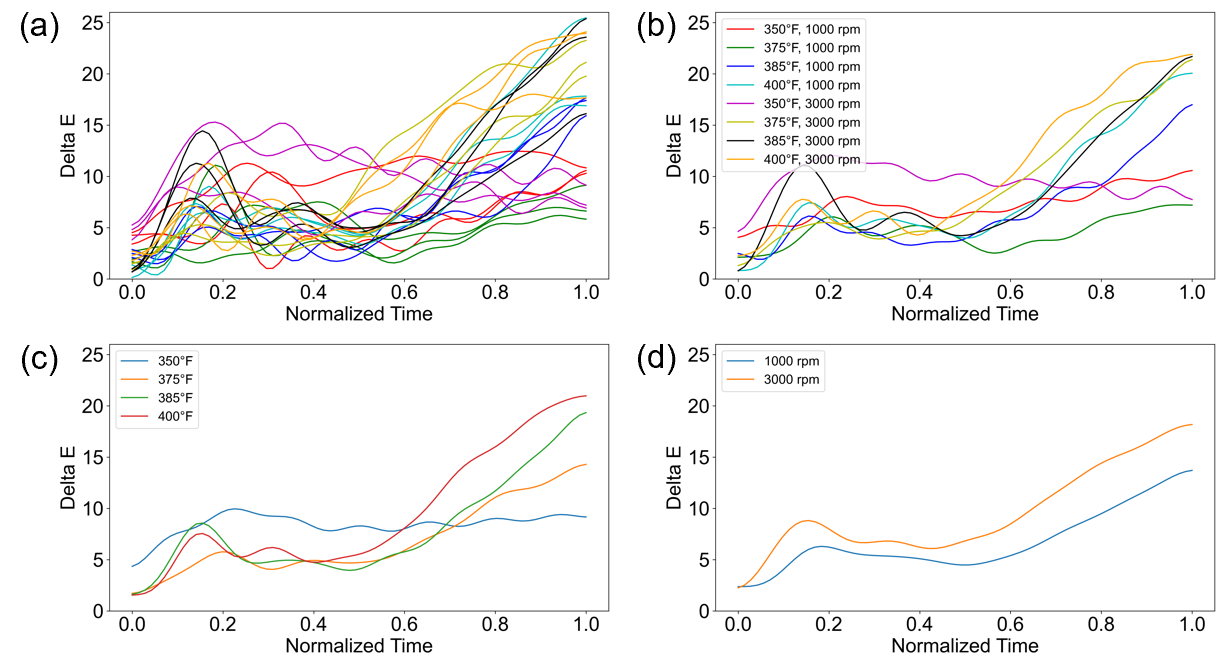}}
\caption{Cookie color trajectories under (a) different drying cases, (b) average trajectory for each process condition, (c) average trajectory for each temperature, and (d) average trajectory for each air velocity.}
\label{fig9}
\end{figure}

Fig.~\ref{fig9} illustrates the smoothed color change trajectories of cookies during drying under various process conditions. In Fig.~\ref{fig9}(a), we plot the trajectories for individual samples across all temperature and air velocity combinations. While there is notable variability between samples, trajectories under the same process parameters tend to cluster together, indicating distinguishable patterns. Fig.~\ref{fig9}(b) presents the average trajectory for each process condition. The differences across curves confirm that color evolution is strongly influenced by drying parameters and thus meaningful to predict.

To further isolate the effects of each process parameter, Fig.~\ref{fig9}(c) shows the average trajectories grouped by temperature, while Fig.~\ref{fig9}(d) shows grouping by air velocity. Both demonstrate clear distinctions in $\Delta E$ trajectories, suggesting that temperature and air velocity have measurable and separable impacts on color change over time.

\subsubsection{Results}

We use RMSE to quantify the average magnitude of prediction errors across the entire color trajectory at each timestamp. RMSE is defined as:

\begin{equation}
\text{RMSE} = \sqrt{ \frac{1}{N} \sum_{i=1}^{N} (\hat{y}_i - y_i)^2 },\label{eq}
\end{equation}
where $\hat{y}_i$ denotes the predicted value at time step $i$, $y_i$ is the corresponding ground truth, and $N$ is the total number of trajectory points. A lower RMSE indicates better alignment between the predicted and actual color trajectories.

Table~\ref{tab:rmse_comparison} summarizes the RMSE values for different model configurations, including a baseline model and multiple ablation variants of our data-driven trajectory model.

\begin{table}[t]
\caption{RMSE \textsc{Comparison Under Different Situations for Cookie}}
\label{tab:rmse_comparison}
\centering
\begin{tabular}{p{3.3cm} p{2.8cm} p{1.1cm} p{3.1cm}}
\toprule
Training dataset & Initial data \newline modalities & RMSE & Improvement \newline compared to row 2 \\
\midrule
All unseen sets       & Baseline model & 31.3591 & - \\
All unseen sets       & Tabular-only   & 3.5912  & - \\
Similarity-informed   & Tabular-only   & 2.7420  & 23.65\% \\
All unseen sets       & Multi-modal    & 2.5723  & 28.37\% \\
Similarity-informed   & Multi-modal    & 2.1170  & 41.05\% \\
\bottomrule
\end{tabular}
\end{table}

The baseline model trained solely on tabular inputs and all unseen training sets performs the worst, highlighting its limitations in modeling long-range temporal dependencies and generalizing across unseen process conditions in zero-shot learning. In contrast, our proposed component function-based model significantly outperforms the baseline model, even when trained on all unseen conditions using only tabular data.

The ablation study yields two key observations:
\begin{itemize}
    \item Multi-modal fusion: Incorporating both tabular and image-derived inputs leads to consistently lower RMSEs compared to tabular-only models, demonstrating the benefit of leveraging richer information for trajectory modeling.
    \item Training set similarity: Using training data from conditions similar to the test set (rather than all unseen conditions) results in better generalization.
\end{itemize}

\subsubsection{Discussions}

\begin{figure*}[!t]
\centerline{\includegraphics[width=\textwidth]{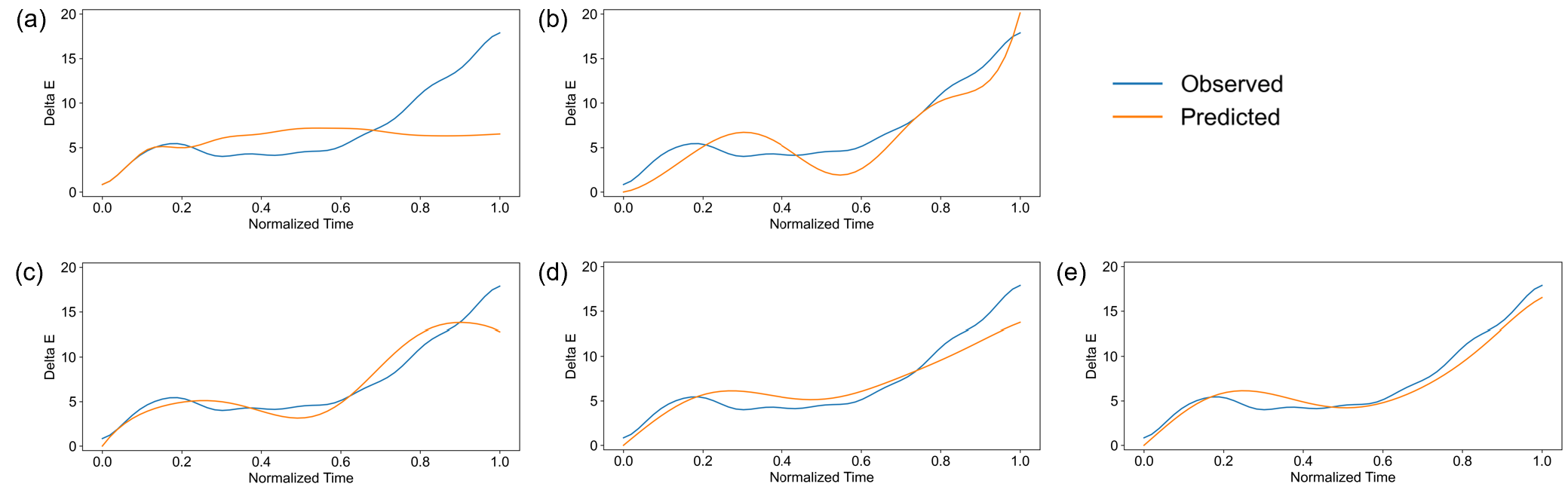}}
\caption{Predicted cookie drying color change trajectories under 400${}^\circ\mathrm{F}$, 1000 RPM for: (a) baseline model, (b) tabular-only model trained on all-unseen set, (c) tabular-only model trained on similarity-informed set, (d) multi-modal model trained on all-unseen set, and (e) multi-modal model trained on similarity-informed set.}
\label{fig10}
\end{figure*}

Fig.~\ref{fig10} presents visual comparisons of predicted and observed color trajectories under a representative test condition (400${}^\circ\mathrm{F}$, 1000 RPM), across five training configurations in a zero-shot setting.

The baseline model (Fig.\ref{fig10}a), trained on all unseen conditions using tabular-only input, exhibits the largest deviation from the ground truth, particularly underestimating the late-stage trajectory. This highlights the limited capacity of process parameters alone to model complex color evolution. Incorporating a similarity-informed training set (Fig.\ref{fig10}c) improves the trajectory alignment, capturing the overall shape more accurately. However, tabular-only models still tend to underpredict the full $\Delta E$ range, especially as the trajectory steepens.

Multi-modal models, which leverage initial-stage visual features, show substantial improvements in prediction quality. As seen in Fig.\ref{fig10}d and \ref{fig10}e, multi-modal training leads to smoother, more accurate predictions that closely follow both early and late-stage color changes. The model trained on the similarity-informed set with multi-modal input (Fig.\ref{fig10}e) achieves the closest match to the ground truth, effectively capturing sharp transitions and nonlinearities in surface appearance.

These results underscore the effectiveness of using visually informative inputs and training with semantically similar conditions. Even without fine-tuning on the target condition, this approach significantly enhances zero-shot generalization for color trajectory prediction.

\subsection{Color Trajectory Prediction in Apple Drying}

\subsubsection{Experiments and Datasets}

Apple drying is a long-duration process. While the color change is less pronounced compared to cookie drying, surface color remains visually trackable and provides meaningful indicators of moisture loss and overall product quality.

We perform controlled drying experiments to collect apple color trajectory data under varying process conditions. We use Fuji apples as the drying sample, which are cored to 2.5cm and sliced transversely to millimeter-level thickness. Slice characteristics such as diameter, thickness, and weight naturally vary, capturing typical process variability.

\begin{figure}[!t]
\centerline{\includegraphics[width=0.85\columnwidth]{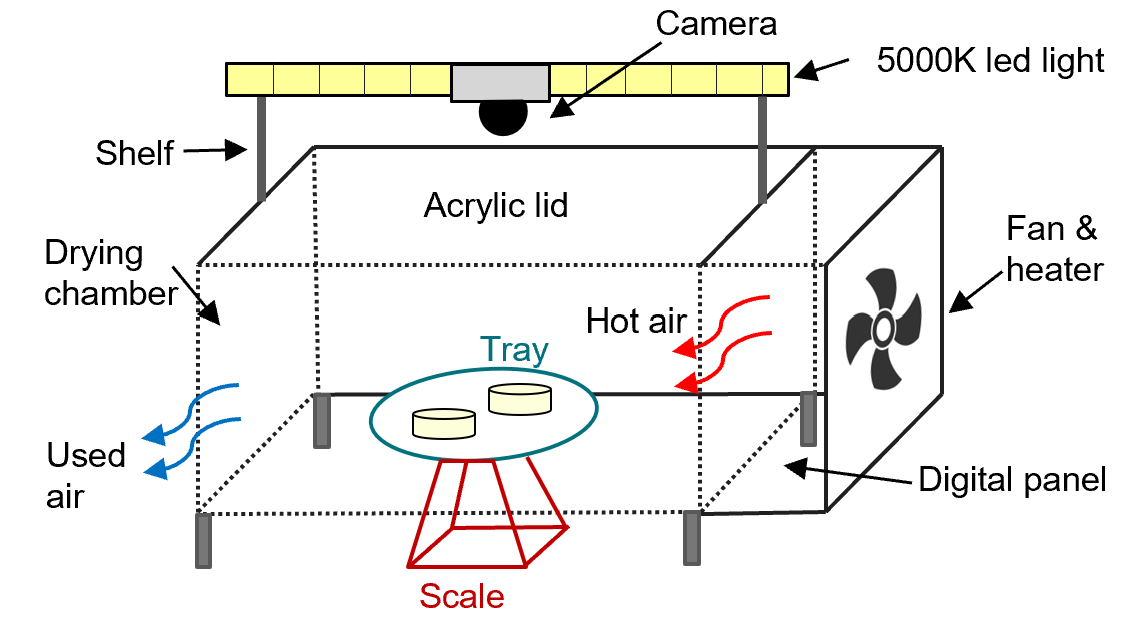}}
\caption{Experimental setup for apple drying.}
\label{fig12}
\end{figure}

Drying is conducted using a custom convective hot-air dryer retrofitted for multi-modal data collection. As shown in Fig.~\ref{fig12}, the system includes a transparent lid, overhead LED lighting, and a camera for in-situ image capture. A scale beneath the tray records weight continuously, allowing real-time moisture tracking.

Experiments span three temperatures (60, 70, 80$\,^{\circ}\mathrm{C}$) and two air velocities (1.5m/s, 2.5m/s). For each setting, samples are dried until reaching one of two target moisture levels (~10\% or ~20\%), with drying durations ranging from 70 to 250 minutes. Each condition includes two repetitions with one slice and three with two slices, totaling 84 samples. 

\begin{figure}[!t]
\centerline{\includegraphics[width=0.8\columnwidth]{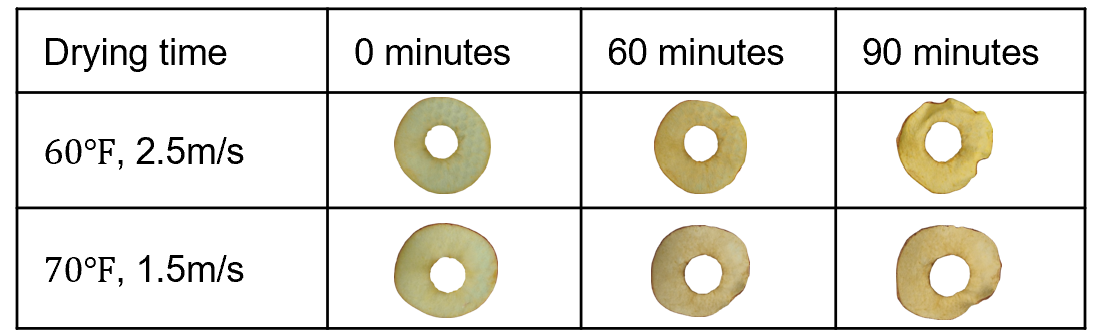}}
\caption{Masked apple samples under 60${}^{\circ}\mathrm{C}$, 2.5m/s and 70${}^{\circ}\mathrm{C}$, 1.5m/s.}
\label{fig13}
\end{figure}

We apply the SAM to extract individual slices in each frame and compute their average surface color over time. Fig.~\ref{fig13} shows an example of two time-series masked apple samples. We extract the pixel-wise average color value from each mask and use it as the trajectory variable over time.

\begin{figure}[!t]
\centerline{\includegraphics[width=0.8\columnwidth]{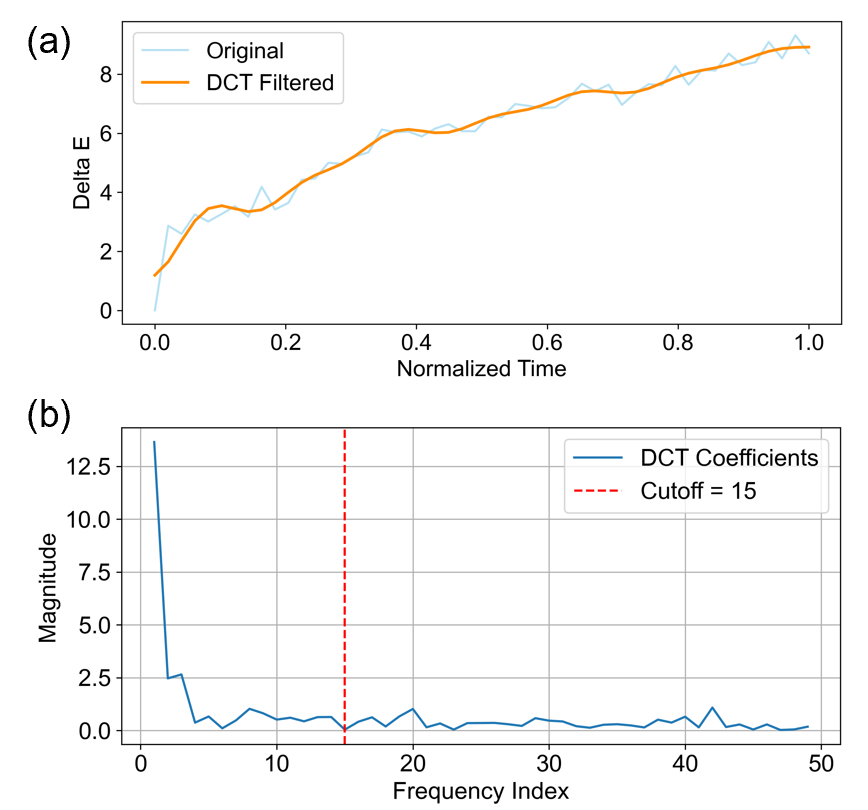}}
\caption{Color trajectory noise cancellation for apple drying with (a) original and smoothed average trajectory under 60${}^{\circ}\mathrm{C}$, 2.5m/s and corresponding (b) DCT coefficient spectrum.}
\label{fig14}
\end{figure}

The resulting $\Delta E$ trajectories are denoised using DCT filtering, with example case shown in Fig.~\ref{fig14}, which preserves the main trend while removing high-frequency noise, with a cutoff threshold at index 15. We validate that both temperature and air velocity have measurable and separable impacts on color change over time for apple drying.

\subsubsection{Results}

To evaluate model performance on the apple drying dataset, we report average RMSEs across process conditions under different training configurations (Table~\ref{tab:rmse2}). The baseline model performs poorly with a high RMSE of 10.1517. Using only tabular data reduces the error significantly, and further improvements are observed by incorporating multi-modal inputs. Training on similarity-informed subsets leads to even better results, with the best configuration—multi-modal inputs and similarity-informed training—achieving an RMSE of 1.2865, a 25.13\% improvement over the tabular-only baseline. These findings demonstrate the model’s robustness and ability to generalize to slower-evolving drying processes.

\begin{table}[t]
\caption{RMSE \textsc{Comparison Under Different Situations for Apple}}
\label{tab:rmse2}
\centering
\begin{tabular}{p{3.3cm} p{2.8cm} p{1.1cm} p{3.1cm}}
\toprule
Training dataset & Initial data \newline modalities & RMSE & Improvement \newline compared to row 2 \\
\midrule
All unseen sets       & Baseline model & 10.1517 & - \\
All unseen sets       & Tabular-only   & 1.7183  & - \\
Similarity-informed   & Tabular-only   & 1.5635  & 9.01\% \\
All unseen sets       & Multi-modal    & 1.3904  & 19.08\% \\
Similarity-informed   & Multi-modal    & 1.2865  & 25.13\% \\
\bottomrule
\end{tabular}
\end{table}

\subsubsection{Discussions}

\begin{figure*}[!t]
\centerline{\includegraphics[width=\textwidth]{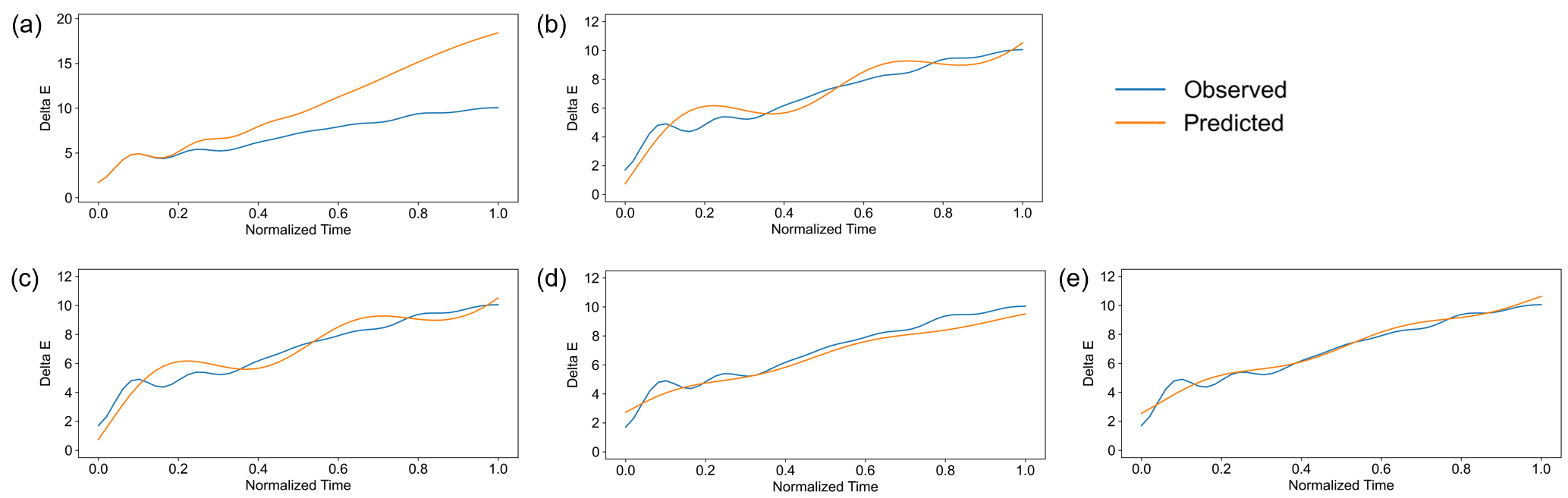}}
\caption{Predicted apple drying color change trajectories under 60${}^\circ\mathrm{C}$, 2.5m/s for: (a) baseline model, (b) tabular-only model trained on all-unseen set, (c) tabular-only model trained on similarity-informed set, (d) multi-modal model trained on all-unseen set, and (e) multi-modal model trained on similarity-informed set.}
\label{fig15}
\end{figure*}

Figure~\ref{fig15} shows predicted versus observed color trajectories for apple drying under a test condition of 60${}^\circ\mathrm{C}$ and 2.5m/s air velocity, comparing models trained under different configurations without target-condition fine-tuning. The tabular-only model trained on all unseen data performs poorly, often misaligning the slope and magnitude of color change. Training on similar-condition subsets improves predictions modestly (Fig.~\ref{fig15}(b)-(c)), indicating better alignment from structurally related data. Incorporating image-based features further enhances prediction quality, with the multi-modal model trained on similar data yielding the most consistent trajectories (Fig.~\ref{fig15}(e)). These results demonstrate that even in gradual color-change scenarios like apples, multi-modal inputs and training set similarity are key to enabling accurate zero-shot generalization.

\section{Conclusion and Future Work}
This work presents a data-driven modeling framework for predicting color change trajectories in food drying processes under unseen conditions. By learning compact representations of color evolution using component functions and incorporating both tabular and image-derived inputs, the proposed approach effectively models dynamic, non-monotonic behavior without requiring condition-specific fine-tuning. We validate the method on cookie and apple drying. The results demonstrate strong zero-shot generalization and consistent performance improvements from multi-modal fusion and training set selection. The method is broadly applicable to other food drying scenarios where surface appearance serves as a proxy for internal quality.

Future work could explore automated clustering of drying regimes or integrate uncertainty-aware modeling to better handle ambiguous or overlapping conditions. Additionally, we plan to extend this framework toward real-time drying control by using predicted trajectories to guide in-situ decision-making, such as early stopping or dynamic temperature adjustments. These extensions would enhance the practical deployment of trajectory-based quality monitoring in industrial food processing systems.


\section*{Acknowledgments}

This study was financially supported by the U.S. Department of Energy, Office of Advanced Manufacturing under Award Number DE-EE0009125, the Massachusetts Clean Energy Center (MassCEC), and Center for Advanced Research in Drying (CARD). The views expressed herein do not necessarily represent the views of the U.S. Department of Energy or the United States Government.


\newpage
\bibliography{mybibfile}

\end{document}